\documentclass{article}

\usepackage{microtype}
\usepackage{graphicx}
\usepackage{subfigure}
\usepackage{booktabs} 

\usepackage[accepted]{icml2019}

\usepackage{wrapfig}
\usepackage{amsmath}
\usepackage{amssymb}
\usepackage{amsthm}
\usepackage{bm}
\usepackage{times}
\usepackage{graphicx} 
\usepackage{subfigure} 
\usepackage{multirow}
\usepackage{color}
\usepackage{algorithm}
\usepackage{algorithmic}
\usepackage{multirow}

\newtheorem{theorem}{Theorem}[section]

\newtheorem{proposition}[theorem]{Proposition}

\icmltitlerunning{Adversarial Self-Paced Learning for Mixture Models of Hawkes Processes}

\begin{document}

\twocolumn[
\icmltitle{Adversarial Self-Paced Learning for Mixture Models of Hawkes Processes}



\icmlsetsymbol{equal}{*}

\begin{icmlauthorlist}
\icmlauthor{Dixin Luo}{du}
\icmlauthor{Hongteng Xu}{in,du}
\icmlauthor{Lawrence Carin}{du}
\end{icmlauthorlist}

\icmlaffiliation{du}{Department of ECE, Duke University, Durham, NC, USA}
\icmlaffiliation{in}{Infinia ML, Inc., Durham, NC, USA}

\icmlcorrespondingauthor{Dixin Luo}{dixin.luo@duke.edu}

\icmlkeywords{Hawkes process, self-paced learning, adversarial learning, mixture model, data augmentation}

\vskip 0.3in
]



\printAffiliationsAndNotice{}  

\begin{abstract}
We propose a novel adversarial learning strategy for mixture models of Hawkes processes, leveraging 
data augmentation techniques of Hawkes process in the framework of self-paced learning. 
Instead of learning a mixture model directly from a set of event sequences drawn from different Hawkes processes, the proposed method learns the target model iteratively, which generates ``easy'' sequences and uses them in an adversarial and self-paced manner. 
In each iteration, we first generate a set of augmented sequences from original observed sequences. 
Based on the fact that an easy sample of the target model can be an adversarial sample of a misspecified model, we apply a maximum likelihood estimation with an adversarial self-paced mechanism. 
In this manner the target model is updated, and the augmented sequences that obey it are employed for the next learning iteration. 
Experimental results show that the proposed method outperforms traditional methods consistently.
\end{abstract}

\section{Introduction}
Real-world event sequences are often modeled based on temporal point processes. 
Specifically, a temporal point process with $C$ event types can be represented as a counting process $N(t)=\{N_c(t)\}_{c\in\mathcal{C}}$, where each $N_c(t)$ is the number of type-$c$ events happening at or before time $t$ and $\mathcal{C}=\{1,...,C\}$. 
As a special kind of point process, Hawkes process~\cite{hawkes1971spectra} formulates the expected instantaneous happening rate of type-$c$ events, or called \emph{intensity function}, as
\begin{eqnarray*}
\begin{aligned}
\lambda_c(t)=\frac{\mathbb{E}[dN_c(t)|\mathcal{H}^\mathcal{C}(t)]}{dt}=\mu_c+\sideset{}{_{i:t_i<t}}\sum\phi_{cc_i}(t-t_i),
\end{aligned}
\end{eqnarray*}
where $\mathcal{H}^\mathcal{C}(t)=\{(t_i, c_i)|t_i<t, c_i\in \mathcal{C}\}$ contains historical events before time $t$, $\mu_c$ is the base intensity capturing exogenous fluctuations of the type-$c$ event, and $\phi_{cc'}(t)$ is the impact function measuring the infectivity of the type-$c'$ event to the type-$c$ event type over time. 
Therefore, we denote an event sequence yielding to a Hawkes process as $\bm{s}\sim \mbox{HP}(\bm{\mu}, \bm{\Phi})$, with basic intensity $\bm{\mu}=[\mu_c]\in\mathbb{R}^{C}$ and impact functions $\bm{\Phi}=[\phi_{cc'}^{k}(t)]$.

As an extension of Hawkes process, 
the mixture model of Hawkes processes (MixHP) is capable of describing the clustering structure of different event sequences and capturing the dependency among events within each cluster. 
MixHP has been used to model real-world event sequences, $e.g.$, patient admissions~\cite{xu2017dirichlet}, social behaviors~\cite{yang2013mixture}, and user logs of information systems~\cite{luo2015multi}. 
Suppose that the sequences in $\mathcal{S}=\{\bm{s}_n=(t_i^n, c_i^n)_{i=1}^{I_n}\}_{n=1}^{N}$  are generated via $K$ different Hawkes processes, $i.e.$, 
\begin{eqnarray}\label{eq:mixHP}
\begin{aligned}
k\sim \bm{\pi},~~\bm{s}|k \sim \mbox{HP}(\bm{\mu}^k, \bm{\Phi}^k),~\text{for}~\bm{s}\in\mathcal{S},
\end{aligned}
\end{eqnarray}
where $\bm{\pi}=[\pi_k]\in\Sigma^{K}$ represents the distribution of the $K$ Hawkes processes. 
Accordingly, the likelihood of a sequence $\bm{s}$ is represented as
\begin{eqnarray}\label{eq:prob}
\begin{aligned}
p(\bm{s};\bm{\Theta}) = \sideset{}{_{k=1}^{K}}\sum \pi_k p(\bm{s}|\bm{\mu}^k, \bm{\Phi}^k).
\end{aligned}
\end{eqnarray}
Here, $p(\bm{s}| \bm{\mu}^k, \bm{\Phi}^k)=\prod_{i=1}^{I}\lambda_{c_i}^k(t_i)e^{-\sum_{c}\int_{0}^{T}\lambda_c^k(s)ds}$ is the likelihood of the sequence $\bm{s}$ conditioned on the $k$-th Hawkes process $\mbox{HP}(\bm{\mu}^k, \bm{\Phi}^k)$.

\begin{figure*}[t]
    \centering
    \subfigure[\tiny{Traditional MLE}]{
    \includegraphics[height=2.3cm]{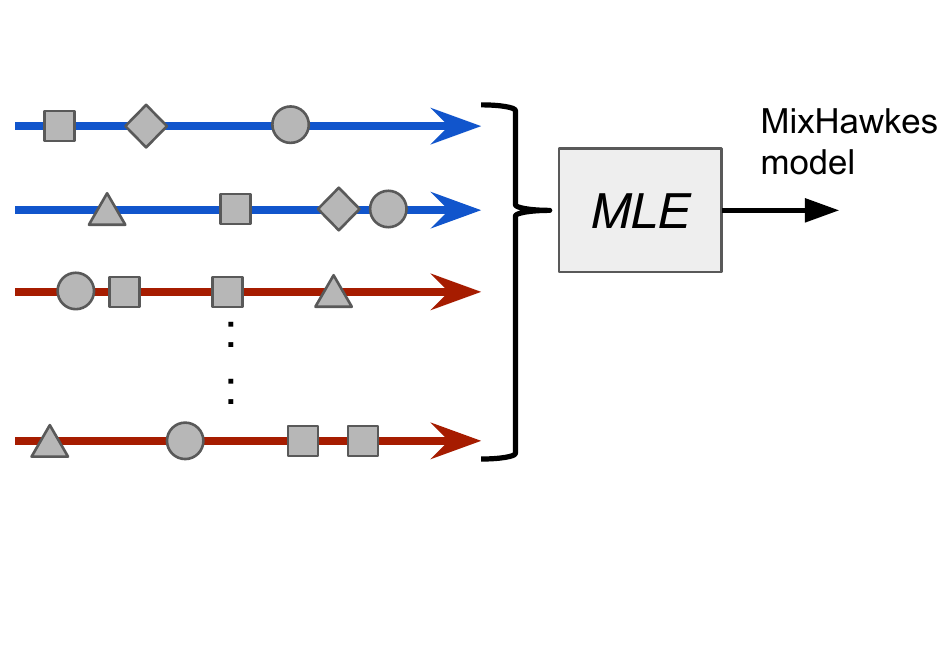}\label{fig:mle}
    }
    \subfigure[\tiny{Self-paced learning}]{
    \includegraphics[height=2.3cm]{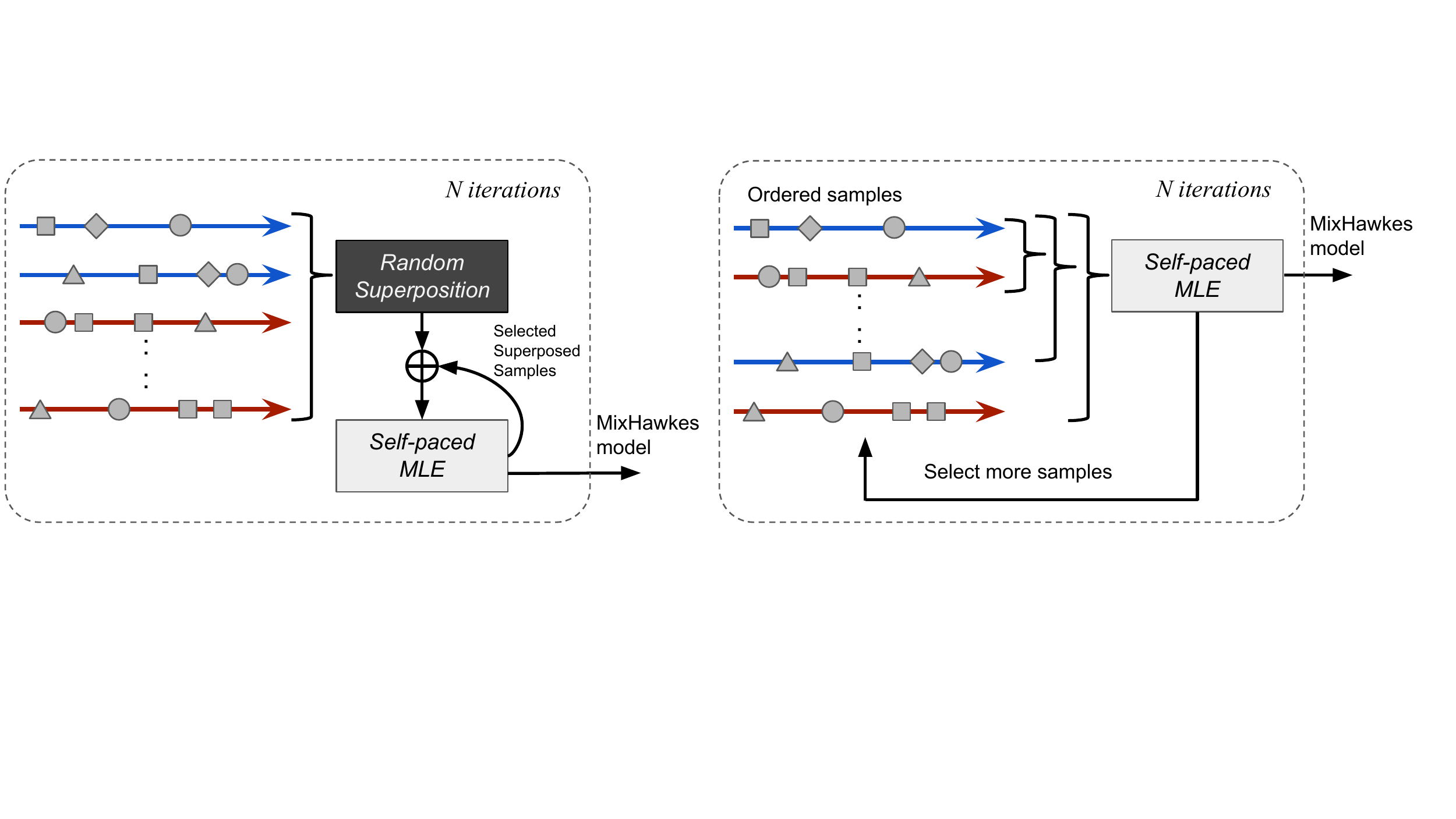}\label{fig:spl}
    }
    \subfigure[\tiny{Adversarial self-paced learning}]{
    \includegraphics[height=2.3cm]{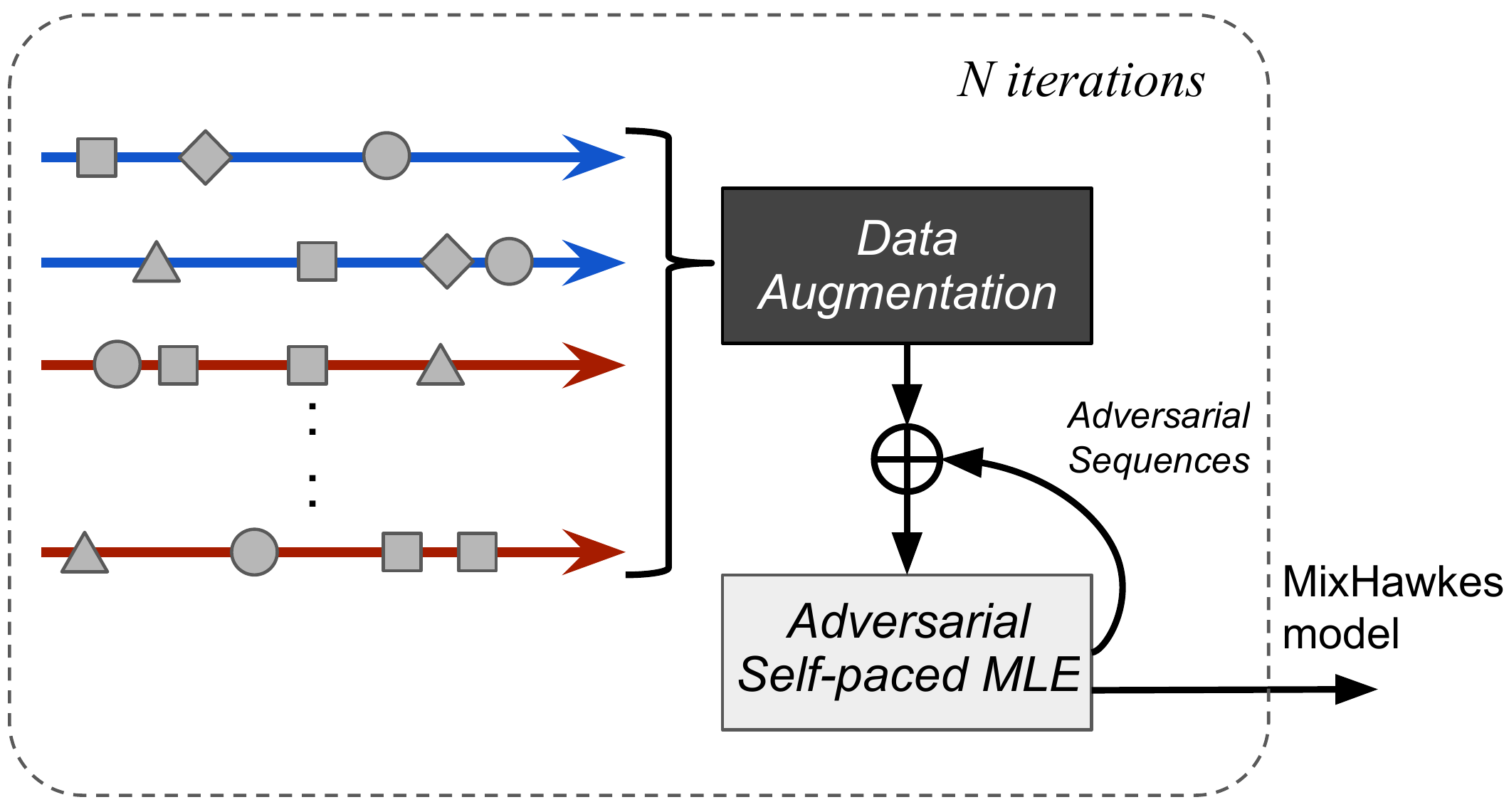}\label{fig:aspl}
    }
    \subfigure[\tiny{Misspecifying models by augmenting data}]{
    \includegraphics[height=2.5cm]{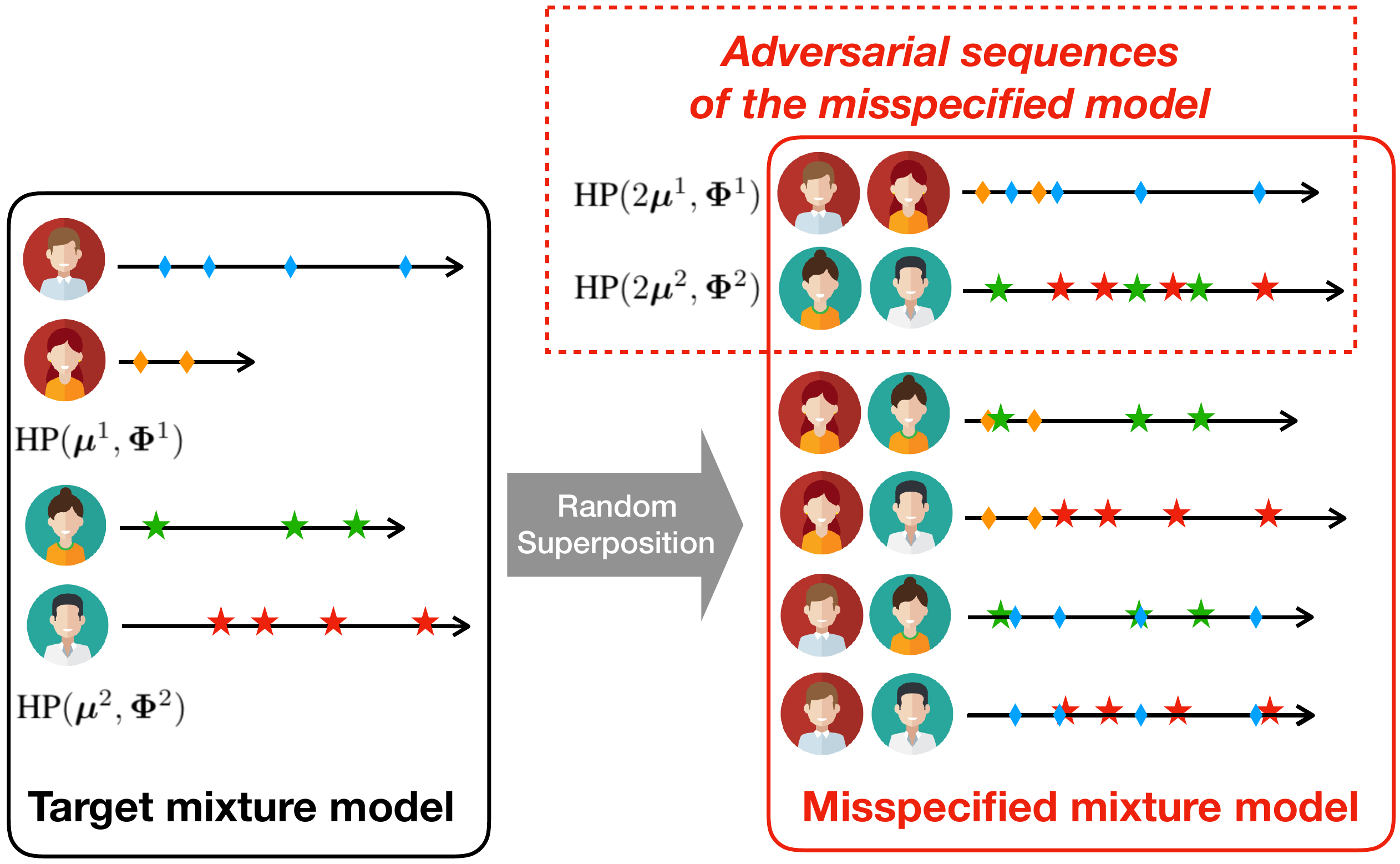}\label{fig:super}}
    \vspace{-10pt}
    \caption{The schemes of various learning methods. The sequences of different Hawkes processes are labeled in different colors.
    }
    \label{fig:scheme}
\end{figure*}

Given observed sequences, we can apply maximum likelihood estimation (MLE) to learn the target mixture model, as shown in Figure~\ref{fig:mle}. 
However, in practice this learning strategy often suffers from insufficienct data. 
For example, in the admission record dataset MIMIC-III~\cite{johnson2016mimic}, most patients only have $5$ admissions or fewer in ten years, while there are over 600 kinds of diseases ($i.e.$, types of events). 
Learning from such short sequences leads to serious over-fitting. 
For a single Hawkes process, such a problem can be mitigated by various data augmentation techniques, $e.g.$, randomly stitching~\cite{xu2017learning} or superposing~\cite{xu2018benefits} the original sequences. 
Unfortunately, for mixture models of Hawkes processes, these techniques cannot be applied directly, because in general the stitching/superposing result of two sequences from different clusters does not yield to a Hawkes process, which may cause serious model misspecification.

To overcome the challenges above, we propose a novel adversarial self-paced learning (ASPL) method and train it iteratively to robustly learn mixture models of Hawkes processes. 
As shown in Figure~\ref{fig:aspl}, in each iteration we actively generate candidates of ``easy'' sequences based on data augmentation methods ($e.g.$, random superposition and stitching). 
Then, based on MLE with an adversarial self-paced~\cite{bengio2009curriculum,kumar2010self} regularizer, we use these candidate sequences to learn the target model and select ``easy'' sequences for the next iteration.

The proposed learning method is based on two facts. 
First, the MixHP model learned from the augmented sequences is always misspecified to some degree, because most of the augmented sequences are not drawn from a Hawkes process. 
As a result, the augmented sequences still obeying Hawkes processes are adversarial samples of the misspecified model.
Second, the easiness of a sample is dependent on the model imposed on it --- an easy sample of the target MixHP model can be an adversarial one of the misspecified model.
Accordingly, our method selects the adversarial sequences of the current misspecified model to construct the easy sequence set for the target model, with this performed in an iterative manner. 
With an increase in iterations, the potential easy sequences become dominant in the training set and the misspecifed model is revised and approaches to the target one.

\section{Adversarial Self-Paced Learning}
\subsection{Data augmentation and model misspecification}\label{sec:data}
For a single Hawkes process, the over-fitting problem caused by insufficient data can be mitigated based on data augmentation techniques. 
In particular, the Hawkes process has an interesting superposition property:
\begin{proposition}[\cite{xu2018benefits}]\label{superpose_th}
Give $M$ independent Hawkes processes with shared impact functions, 
$i.e.$, $\mbox{HP}(\bm{\mu}^m, \bm{\Phi})$ and $N^m(t)\sim \mbox{HP}(\bm{\mu}^m, \bm{\Phi})$ for $m = 1, ..., M$, 
their superposition becomes a single Hawkes process, 
$i.e.$, $N(t)=\sum_{m=1}^M N^m(t)$ and $N(t)\sim \mbox{HP}(\sum^M_{m=1}\bm{\mu}^m, \bm{\Phi})$.
\end{proposition}
Additionally, for a stationary Hawkes process, its impact function $\phi_{cc'}(t)$ satisfies $\int_{0}^{+\infty}\phi_{cc'}(s)ds<+\infty$, implying that the infectivity of a historical event to current one decays rapidly with respect to the time interval between them, $i.e.$, $\lim_{t\rightarrow +\infty}\phi_{cc'}(t)=0$. 
Therefore, given two short sequences belonging to the same Hawkes process, superposing or stitching them can generate a denser or longer sequence for the target Hawkes process model.
These two data augmentation strategies have been applied to learn Hawkes processes from imperfect observations~\cite{xu2018benefits,xu2017learning,xu2018learning}, which indeed improve learning results.

However, as shown in Figure~\ref{fig:super}, when the sequences generated by different Hawkes processes with different impact functions, their superposition/stitching result  will not yield a Hawkes process any more. 
Therefore, most of the augmented sequences do not obey the target mixture model of Hawkes processes, learning from which leads to a \emph{misspecified MixHP model}, 
while those really obeying Hawkes processes are the minority of the augmented sequences, will be ignored and treated as \emph{adversarial samples}~\cite{lowd2005adversarial,barreno2006can,liu2009game,huang2011adversarial} of the misspecified model. 

\subsection{The easiness of sequence}
Although directly applying traditional data augmentation techniques ($i.e.$, superposing and stitching) is not helpful to learn mixture models of Hawkes processes, the augmented sequences have different levels of fitness with respect to the misspecified model, which provides a reasonable measurement for the easiness of the sequences, and makes self-paced learning possible. 
Specifically, the likelihood of a sequence under a model reflects the fitness of the model to the sequence. 
Given a sequence $\bm{s}$ with $I$ events, we define the easiness of a sequence with respect to the model $\bm{\Theta}=\{\bm{\mu}_k,\bm{\Phi}_k\}_{k=1}^{K}$: 
\begin{eqnarray}\label{eq:easy2}
\begin{aligned}
\mbox{easiness}(\bm{s};\bm{\Theta}) = \max_{k\in\{1,..,K\}}\frac{1}{I}\log p(\bm{s}|\bm{\mu}^k,\bm{\Phi}^k).
\end{aligned}
\end{eqnarray}
(\ref{eq:easy2}) indicates that an easy sample of a mixture model needs to fit one of the clustering component with high probability (even if the probability of the component itself is low). 
The higher the likelihood is, the easier the sequence is under the given model.
Dividing by $I$, the easiness of the sequences with different lengths becomes comparable. 
Because (\ref{eq:easy2}) is not differentiable, in practice we can use ``LogSumExp'' operation to achieve a smooth maximum. 
Accordingly, (\ref{eq:easy2}) can be rewritten as
\begin{eqnarray}\label{eq:easy3}
\begin{aligned}
\mbox{easiness}(\bm{s};\bm{\Theta}) = \frac{1}{I}\log\Bigl(\sideset{}{_{k=1}^{K}}\sum p(\bm{s}|\bm{\mu}^k,\bm{\Phi}^k)\Bigr).
\end{aligned}
\end{eqnarray}
In this case, we define adversarial sequences of our model as those with lowest easiness.

\subsection{Proposed learning algorithm}
The key idea of our learning method is that the easy sequences of the target mixture model can be the adversarial ones of the current estimated model. 
When learning a mixture model with potential risk of misspecification based on augmented sequences, we want to find its adversarial sequences and add them into the ``easy'' sequence set of the target mixture model. 
The easy sequences are considered in the next training iteration, which are used to revise the misspecified model.

In the $m$-th learning iteration, given the augmented sequences $\mathcal{S}^{(m)}$ and the easy sequence set $\mathcal{S}_{easy}$ generated in the previous iteration, we update the current mixture model and select new easy sequences from $\mathcal{S}^{(m)}$ simultaneously, by solving the following max-min optimization problem.
\begin{eqnarray}\label{eq:maxmin}
\begin{aligned}
&\max_{\bm{\Theta}\geq\bm{0}}\min_{\bm{w}\in\{0,1\}^{|\mathcal{S}^{(m)}|}}\sideset{}{_{\bm{s}\in \mathcal{S}^{(m)}\cup\mathcal{S}_{easy}}}\sum\underbrace{\log p(\bm{s};\bm{\Theta})}_{\text{log-likelihood}} \\
&\quad +\alpha\sideset{}{_{\bm{s}_n\in\mathcal{S}^{(m)}}}\sum\Bigl[\underbrace{w_n\mbox{easiness}(\bm{s}_n;\bm{\Theta})+ \zeta(1-w_n)}_{\text{adversarial self-paced regularizer}}\Bigr].
\end{aligned}
\end{eqnarray}
Here, $\bm{w}=[w_n]\in\{0,1\}^{|\mathcal{S}^{(m)}|}$ is a binary vector, whose element $w_n$ indicates whether $\bm{s}_{n}$ is an easy sequence of the proposed model. 
The first term represents the log-likelihood of the current model given the whole sequence set, while the second term is the proposed adversarial self-paced regularizer, that measures the easiness of each $\bm{s}_n\in \mathcal{S}^{(m)}$ and selects the adversarial sequences of the current model as the easy sequences of the target model. 
Hyperparameter $\alpha$ controls the significance of the proposed regularizer, and $\zeta$ controls the acceptance rate of easy sequences. 

\begin{algorithm}[t]
	\caption{Adversarial Self-Paced Learning for MixHP}
	\label{alg1}
	\begin{algorithmic}[1]
		\STATE \textbf{Input:} Original sequences $\mathcal{S}=\{\bm{s}_n\}_{n=1}^{N}$.
		\STATE \textbf{Output:} Parameters $\widehat{\bm{\Theta}}$.
		\STATE Initialize easy sequence set $\mathcal{S}_{easy}=\emptyset$. Set $m=0$.
		\WHILE{$|\mathcal{S}_{easy}|< 2N$}
		\STATE Augment original sequences and get $\mathcal{S}^{(m)}$. 
		\STATE Initialize $\bm{w}=\bm{0}$  
		\WHILE{Not converge}
		\STATE Update $\widehat{\bm{\Theta}}$ via solving (\ref{eq:aspl1}), and set $\zeta$ accordingly. 
		\STATE Given $\zeta$, update $\bm{w}$ via selecting $L$ sequences $\mathcal{S}_{select}$ with lowest easiness.
		\ENDWHILE
		\STATE Update easy sequence set: $\mathcal{S}_{easy}=\mathcal{S}_{easy}\cup\mathcal{S}_{select}$.
		\STATE $m=m+1$.
		\ENDWHILE
	\end{algorithmic}
\end{algorithm}

We decompose (\ref{eq:maxmin}) into two sub-problems and solve them via alternating optimization. 
In each learning iteration, we solve the following two sub-problems:

1) Update current model:
\begin{eqnarray}\label{eq:aspl1}
\begin{aligned}
\widehat{\bm{\Theta}}=\arg&\sideset{}{_{\bm{\Theta}\geq\bm{0}}}\max\sideset{}{_{\bm{s}\in \mathcal{S}^{(m)}\cup\mathcal{S}_{easy}}}\sum\log p(\bm{s};\bm{\Theta}) \\
&+\alpha\sideset{}{_{\bm{s}_n\in\mathcal{S}^{(m)}}}\sum\hat{w}_n\mbox{easiness}(\bm{s}_n;\bm{\Theta}),
\end{aligned}
\end{eqnarray}
where $\hat{w}_n$ is the indicator learned in the previous step.

2) Select new easy sequences for target model:
\begin{eqnarray}\label{eq:aspl2}
\begin{aligned}
\hat{\bm{w}}=\arg\sideset{}{_{\bm{w}\in\{0,1\}^{|\mathcal{S}^{(m)}|}}}\min&\sideset{}{_{\bm{s}_n\in\mathcal{S}^{(m)}}}\sum\Bigl[\zeta(1-w_n) \\
&+w_n\mbox{easiness}(\bm{s}_n;\widehat{\bm{\Theta}}) \Bigr].
\end{aligned}
\end{eqnarray}

Maximizing the easiness term in (\ref{eq:aspl1}) makes the model fit the selected easy sequences and suppresses the influence of those ``non-Hawkes'' sequences. 
When selecting new easy sequences, on the contrary, we keep the sequences with low easiness with respect to current model for the following learning iterations. 
Algorithm~\ref{alg1} shows the scheme of the proposed learning method.

As mentioned in line 8 of Algorithm~\ref{alg1}, we set $\zeta$ according to the learning result of (\ref{eq:aspl1}). 
Given current mixture model $\widehat{\bm{\Theta}}$ and learned distribution of clustering component $\hat{\bm{\pi}}=[\hat{\pi}_k]$, for all $\bm{s}_n\in\bm{S}^{(m)}$, we first sort $\mbox{easiness}(\bm{s}_{n};\widehat{\bm{\Theta}})$ in ascending order, and then select the top-$L$ augmented sequences.
Because the proportion of adversarial sequences in $\bm{S}^{(m)}$ can be approximated as $\sum_{k}\pi_k^2$, the number of easy sequences should not be larger than $\sum_{k}\pi_k^2|\bm{S}^{(m)}|$.
We use $\hat{\pi}_k$ to estimate the real $\pi_k$ and set $L=\lfloor 0.25\sum_{k}\hat{\pi}_k^2|\bm{S}^{(m)}|\rfloor$. 
Accordingly, $\zeta = \mbox{easiness}(\bm{s}_{L};\widehat{\bm{\Theta}})$, 
where $\bm{s}_{L}$ is the $L$-th sorted sequence. 

\begin{table*}[t]
\caption{Comparisons for various methods on real-world data.}
\centering
\small{
\setlength{\tabcolsep}{5pt}
\begin{tabular}{c|
c@{\hspace{5pt}}
c@{\hspace{5pt}}
c@{\hspace{5pt}}
c|
c@{\hspace{9pt}}
c@{\hspace{9pt}}
c@{\hspace{9pt}}
c@{\hspace{5pt}}
c@{\hspace{5pt}}
c} 
\hline\hline
\multirow{3}{*}{Dataset}
&\multicolumn{4}{c|}{\multirow{2}{*}{Setting}} 
&\multirow{2}{*}{MMHP} 
&\multirow{2}{*}{DMHP} 
&DMHP
&\multirow{2}{*}{SPL-MixHP} 
&ASPL-MixHP 
&ASPL-MixHP\\
& & & & &&&Stitch & &Stitch &Superpose\\ \cline{2-11}
&$N_{train}/N_{test}$ &$T$ &$C$ &$K$ &loglike &loglike &loglike &loglike &loglike &loglike \\
\hline
MIMIC-III &903 / 226 &10 yrs &8 &10 
&-3.46$\pm$0.71 
&-2.85$\pm$0.29 
&-2.90$\pm$0.20
&-2.66$\pm$0.12 
&-2.24$\pm$0.10 
&\textbf{-2.07$\pm$0.08} \\
IPTV &15103 / 15103 &24 hrs &16 &10 
&-0.53$\pm$0.13 
&1.38$\pm$0.11
&1.25$\pm$0.07
&1.37$\pm$0.09 
&\textbf{1.45$\pm$0.03} 
&1.44$\pm$0.02 \\
LinkedIn &1220 / 1219 &15 yrs &82 &5
&-7.39$\pm$0.33 
&-4.69$\pm$0.20
&-4.92$\pm$0.14
&-4.64$\pm$0.16 
&\textbf{-3.97$\pm$0.12} 
&-4.02$\pm$0.14 \\
\hline\hline
\end{tabular}\label{tab:real}
}\vspace{-5pt}
\end{table*}

\subsection{Complexity}
Given $N$ sequences with $I$ events per each, 
the computational complexity for learning a mixture model of $K$ Hawkes processes is $\mathcal{O}(KNI^2)$. 
Applying the proposed learning strategy, we need to update the model based on various augmented sequence sets in different iterations, and each augmented sequence may have $2I$ events. 
Denote the maximum number of iterations as $M$. 
The computational complexity of our method is $\mathcal{O}(4MKNI^2)$. 
Fortunately, the proposed learning method is mainly designed for the case of short sequences, whose numbers of events are often very small. 
Given the improvements on learning results brought from our learning method, which will be shown in the following section, the increase of the computational complexity appears to be tolerable. 


\section{Experiments}
We denote our adversarial self-paced learning method for mixture models of Hawkes processes as \textbf{ASPL-MixHP}. To demonstrate its effectiveness, we compare our method with state-of-the-art methods on three real-world datasets. 
In particular, we consider four competitive alternatives to our method. 
$i$) \textbf{MMHP}: The multi-task multi-dimensional Hawkes process~\cite{luo2015multi}, which learns one Hawkes process per sequence and applies $K$-means to the learned clusters of all sequences.
$ii$) \textbf{DMHP}: The Dirichlet mixture model of Hawkes processes~\cite{xu2017dirichlet}, which learns the proposed mixture model directly from observed sequences based on variational inference. 
$iii$) \textbf{DMHP-Stitch}: The DMHP model learned based on the augmented sequences generated by random stitching.
$iv$) \textbf{SPL-MixHP}: The self-paced learning of the mixture model of Hawkes process, which applies the original self-paced learning strategy~\cite{kumar2010self}, $i.e.$, in each iteration, we select the sequences with the highest likelihood per event for the next learning iteration, to learn the target mixture model via MLE, as shown in Figure~\ref{fig:spl}. 
For our ASPL-MixHP method, both random superposition and random stitching are applied as feasible data augmentation methods. 
The hyperparameter $\alpha$ is set to be $10$ empirically in the following experiments. 

After learning models based on $N_{train}$ training sequences, we evaluate the performance of various methods on $N_{test}$ testing sequences, calculating the average log-likelihood of the testing sequences
\begin{eqnarray}
\begin{aligned}
\text{loglike}=\frac{1}{N_{test}}\sideset{}{_{n=1}^{N_{test}}}\sum\frac{1}{I_n}p(\bm{s}_n;\widehat{\Theta}).
\end{aligned}
\end{eqnarray}
This measurement reflects the fitness of a trained model to the testing samples.
Each method is tested in 15 trials.
In each trial, the sequences are randomly divided into training and testing sets. 
The model trained on the training set is applied to the testing set. 
The averaged testing log-likelihood and its 95\% confidence interval are calculated. 

We apply our method to $i$) cluster LinkedIn users according to their job-hopping behaviors~\cite{xu2018learning}, 
$ii$) cluster patients according to their admissions~\cite{xu2017learning}, 
and $iii$) cluster IPTV users according to their daily viewing records~\cite{luo2014you}.
These three datasets suffer from data sparsity --- generally, each sequence in these three datasets contains just $10$ events or fewer. 
For the LinkedIn dataset, in each trial the job-hopping behaviors of $1220$ LinkedIn users are used to train a mixture model, and the records of the remaining $1219$ users are used for testing. 
These records involve $82$ IT companies and universities, which are treated as the event types in the model. 
For the MIMIC-III dataset, the diseases in patients' admissions are categorized into $8$ classes.
For each patient, his/her admissions in ten years are observed event sequences, which are modeled as a mixture model of $10$ Hawkes processes. 
We use $903$ sequences for training and $226$ sequences for testing in each trial.
For the IPTV viewing records, we obtain $15103$ daily viewing records of $16$ kinds of TV programs from $1000$ users in each trial, for training a mixture model of $10$ Hawkes processes. 
The records of the following days are used for testing the model.

Table~\ref{tab:real} lists the results of various methods on three real-world datasets. 
Experimental results show that our ASPL-MixHP method works well, obtaining higher testing log-likelihood than the other methods.

\section{Conclusion and Future Work}
We propose an adversarial self-paced learning method for mixture models of Hawkes processes. 
Our method combines data augmentation techniques with a self-paced learning strategy, generating and selecting easy sequences iteratively for the target model, from the adversarial sequences of a potentially misspecified model. 
We test our method on real-world datasets and demonstrate its potential to improve learning results in cases with short training sequences.
In the future, we plan to further reduce its computational complexity and improve its scalability to imblanced large-scale clustering problems. 
Beyond mixture models of Hawkes processes, we will extend the proposed method to the mixture models of other temporal point processes.

\textbf{Acknowledgments} 
This research was supported in part by DARPA, DOE, NIH, ONR and NSF. 

\newpage
\bibliographystyle{icml2019}
\bibliography{aspl}

\end{document}